\pdfoutput=1

\documentclass[11pt]{article}

\usepackage{acl}
\usepackage{booktabs}
\usepackage{times}
\usepackage{latexsym}

\usepackage[T1]{fontenc}

\usepackage[utf8]{inputenc}
\usepackage{float}
\usepackage{microtype}
\usepackage{graphicx}
\title{Seamlessly Integrating Factual Information and Social Content with Persuasive Dialogue}

\author{Maximillian Chen$^1$, 
  Weiyan Shi$^{1}$, 
  Feifan Yan$^{1}$,
  Ryan Hou$^1$, \\
  \textbf{Jingwen Zhang}$^2$, 
  \textbf{Saurav Sahay}$^3$, 
  \textbf{Zhou Yu}$^1$\\
  $^1$Columbia University, New York, NY \\
  $^2$University of California, Davis, Davis, CA, $^3$Intel Labs \\
  \texttt{maxchen@cs.columbia.edu} \\
  \texttt{\{ws2634, fy2241, rh2920, zy2641\}@columbia.edu} \\
  \texttt{jwzzhang@ucdavis.edu}, \texttt{saurav.sahay@intel.com}
}
\begin{document}
\maketitle
\begin{abstract}
Complex conversation settings such as persuasion involve communicating changes in attitude or behavior, so users' perspectives need to be  addressed, even when not directly related to the topic.  In this work, we contribute a novel modular dialogue system framework that seamlessly integrates factual information and social content into persuasive dialogue.
Our framework is generalizable to any dialogue tasks that have mixed social and task contents. We conducted a study that compared user evaluations of our framework versus a baseline end-to-end generation model. We found our framework was evaluated more favorably in all dimensions including competence and friendliness, compared to the end-to-end model which does not explicitly handle social content or factual questions. 
\end{abstract}

\newcommand{\persuasion}{\textsc{PersuasionForGood }}
\newcommand{\framework}{\textsc{RAP }}
\newcommand{\frameworknospace}{\textsc{RAP}}

\section{Introduction}
Persuasive dialogue systems are designed for chatbots to communicate with and to influence users with specific goals. Such systems are often designed to benefit individual users (e.g., promoting healthy behaviors) or society at large (e.g., persuading people to make donations). \citet{wang2019persuasion} introduced this idea with the \persuasion dataset, which contains 1,017 human-human conversations where one participant persuaded the other to donate to the charitable organization \textit{Save the Children\footnote{https://www.savethechildren.net/}}, with 300 conversations having sentence-level dialogue act annotations.

The social and communicative dynamics behind persuasive conversation contexts are complex. A persuasive conversation by definition involves one party, the persuader, intending to change the attitude or behavior of the other party, the persuadee~\cite{torning2009persuasive}. Changing persuadees' attitude has several dimensions including establishing mutual trust and credibility, strategically presenting persuasive appeals, and eliciting emotional reactions from the persuadee~\cite{o2015persuasion,wilson2003perceived}.
\begin{figure}
    \centering
    \vspace{10pt}
    \includegraphics[width=0.9\linewidth]{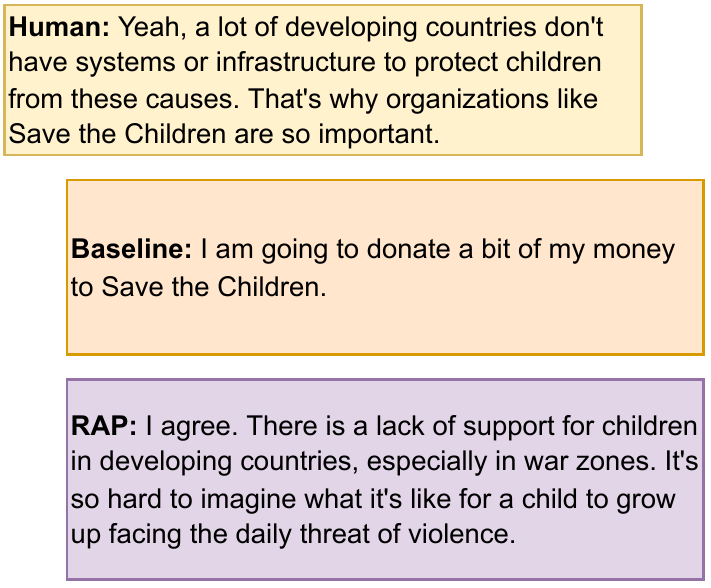}
    \caption{Chatbot running on the baseline BART model and chatbot running on RAP responding to the same user utterance. The baseline model does not appropriately acknowledge the user's statement, whereas RAP is able to show acknowledgement and respond appropriately.}
    \vspace{-10pt}
    \label{fig:comparison}
\end{figure}
Moreover, Grice's Maxims of Conversation define conversations as a cooperative and collaborative process~\cite{grice1975logic,clark1996using,merrison2002politeness}. Thus, effective and successful persuasive conversations do not mechanically relay task-related information to the persuadee. There has to be a significant exchange of social and emotional content to empathetically address persuadees, e.g. by answering specific questions and developing positive relationships throughout the conversation. 

For this reason, persuasive conversations are not strictly task-oriented, but are built around tasks with additional social conversational strategies. In essence, persuasive conversations have two goals: one that is task-oriented to elicit behavioral changes, and another that is social-oriented to build trust and empathy and develop positive relationships in order to better navigate the persuasive context. 
In this work, we propose the Response-Agenda Pushing Framework (\textsc{RAP}) for persuasive dialogue systems, which can explicitly handle these two goals.  
\framework jointly addresses social response and task-oriented dialogue generation. In a given turn, \framework first focuses ond appropriately triggering modules to generate answers to factual questions and social responses to address users' comments. \framework then pushes the persuasive agenda of a conversation using a language model that conditions on individual persuasive appeals.
Compared to state-of-the-art end-to-end conditional generation models, \framework is more semantically coherent and persuasive, while being generalizable to any dataset annotated with dialogue acts. In addition, we tackle the challenge of multiple-sentence conditional generation in a single turn given specific pragmatic argumentative strategies (e.g., ``emotional appeal'').  

Concretely, our contributions are threefold. Contrary to recent work which attempts to transition from social to task-oriented dialogue~\cite{chiu2022salesbot}, we blend social and task-oriented dialogue in an approach grounded in social science theory postulating the need for social acknowledgement in the midst of advancing conversational goals~\cite{o2015persuasion,wilson2003perceived,zhang-danescu-niculescu-mizil-2020-balancing,grice1975logic,merrison2002politeness}. Additionally, we present an account of conditional generation on fine-grained pragmatic persuasive strategies, unlike earlier attempts using looser semantic controls~\cite{he2018decoupling,lewis-etal-2017-deal, hua2019sentence}. Finally, we present a qualitative account of \textsc{RAP}, including individual anecdotes of its strengths and weaknesses. Overall, we present a novel perspective on persuasive dialogue, marking important progress towards intelligent persuasive agents.

\section{Related Work}
Much earlier work in persuasion-like social conversations has been towards building dialogue systems for negotiation tasks, e.g., using the Craigslist Bargaining~\cite{he2018decoupling} and Deal or No Deal datasets~\cite{lewis-etal-2017-deal}. However, in negotiation tasks, the goal is to come to a consensus, whereas in persuasion tasks, the target result is a one-sided change or a ``win'' for the persuader, as in a debate. Recently, there has been increasing interest in persuasive dialogues because of the rise in online-mediated persuasion scenarios (e.g. online sales, health promotion, political debates); much work focuses on understanding the social dynamics behind online persuasive conversations on social media platforms like Reddit (e.g.~\citet{atkinson-etal-2019-gets, musi2018did, srinivasan2019content, tan2016winning}). In addition, a burgeoning line of work has been invested in developing chatbots to deliver healthcare remotely and to persuade people to adopt healthier lifestyles~\cite{oh2021systematic,zhang2020artificial}. Such efforts have inspired a growing body of work towards building persuasive dialogue systems that are \textit{conditional, strategic and factual} to benefit individuals and society at large. 

\begin{table*}[]
\small
\centering
\begin{tabular}{ll}
\toprule
\textbf{Dialogue Act/Persuasive Strategy} & \textbf{Example Utterance} \\\midrule
Greeting & Hello there! How are you doing? \\\hline 
Source-related inquiry & Have you heard of the organization Save the Children? \\\hline  
Personal-related inquiry & Do you have kids yourself? \\\hline  
Credibility appeal & \begin{tabular}[c]{@{}l@{}}Save the Children is an international non-governmental organization that \\ promotes children's rights, provides relief, and helps support children in \\ developing countries.\end{tabular} \\\hline  
Emotional appeal & \begin{tabular}[c]{@{}l@{}}It make me feel sad to see that so many children are suffering from \\ poverty and hunger.\end{tabular} \\ \hline 
Logical appeal & \begin{tabular}[c]{@{}l@{}}Donations are extremely important in order for children to have their \\ rights to healthcare, education, safety, etc. If you were to donate, you \\ would be making a huge impact on these children and on the world.\end{tabular} \\ \hline 
Self-modeling & I think I'll donate a bit of my money to Save the Children, \$2. \\\hline 
Foot-in-the-door & Every little bit helps. Even a small amount! \\ \hline 
Personal story & \begin{tabular}[c]{@{}l@{}}Someone told me that he and his brother replaced birthday gifts \\ with charity donations a few years ago, and it was a really rewarding \\ experience for them.\end{tabular} \\ \hline 
Propose donation & Would you like to make a donation to Save the Children? \\ \hline 
Closing & Thank you, it's been lovely talking to you. Enjoy your day and bye! \\ \hline 
\end{tabular}
\caption{Examples of each dialogue act from \persuasion used for the chatbot.}
\label{dialogue_acts}
\end{table*}
Many early iterations of persuasive dialogue systems have used template-based \cite{zhao2018sogo} or retrieval-based \cite{hiraoka2015evaluation,yoshino-etal-2018-dialogue} utterance generation methods. \citet{wang2019persuasion} introduced \persuasion and proposed designing a personalized persuasive dialogue system. \citet{wu-etal-2021-alternating} used two pre-trained language models to separately models both speakers in a conversation, finding success in creating human-like utterances without supervision (from human annotations). Other studies propose end-to-end neural generation models~\cite{li2020end,lewis-etal-2017-deal}. However, in approaches solely performing language modeling, there is less semantic control over generated utterances; they are not guaranteed to follow a particular persuasive strategy or dialogue act. Beyond persuasion, conditional text generation has emerged as a popular method of controllable generation for more coherent and ``harmonious'' human-dialogue system interactions \cite{guo2021conditional,keskar2019ctrl}. Much earlier work in sentence-level conditional text generation has facilitated control by conditioning on entire topic statements \cite{hua2019sentence} or simple semantic codes \cite{keskar2019ctrl,he2018decoupling,see2019makes}. While such approaches work well in chit-chat, they do not guarantee strategy execution for complex tasks. \textit{We propose using conditional generation conditioned on  pragmatic dialogue acts to specifically control the strategic flow of a persuasive conversation.}

Much existing work in persuasion tasks has focused on strategy/policy planning \cite{georgila2011reinforcement, sakai2020hierarchical,hiraoka2014reinforcement,hiraoka2013dialogue,tran2022ask,black2014automated}, while others have focused on classification~\citet{CHEN202147,tian2020understanding,wang2019persuasion}. Other work discussed challenges in building dialogue systems that are social in nature, stating that unlike task-oriented dialogue systems, open-domain social dialogue systems should form a consistent personality to develop users' trust, satisfy the human need for affection and social belonging, and generate interpersonal responses \cite{huang2020challenges,zhou2020design,walker2004generation} suitable for any input~\cite{higashinaka2014towards}. Consistent with this need for affection and acknowledgement, \citet{zhang-danescu-niculescu-mizil-2020-balancing} find that in crisis counseling, it is necessary to balance the goals of both ``empathetically addressing the crisis situation'' and ``advancing the conversation towards a resolution.'' Additionally, \citet{sun-etal-2021-adding} improved engagement with task-oriented dialogues by adding ``chit-chat.'' This suggests that balancing the need for human acknowledgement with advancing towards conversational goals may improve persuasion outcomes. Very recent work has made progress by transitioning from chit-chat to task-oriented dialogue~\cite{chiu2022salesbot}. \textit{However, to truly achieve this balance, we propose interweaving social content with pushing a conversational agenda in order to improve coherence, friendliness, and persuasiveness.}

Retrieval-based dialogue systems have long been considered one of the core classes of conversational systems~\cite{banchs2012iris}, often being used for question answering systems~\cite{gao2019neural} due to their ability to return ``fluent and informative responses'' \cite{yang2019hybrid}. But, recent work has been able to directly improve their open-domain dialogue systems by ensembling both retrieval methods (e.g., database queries) with neural generation methods \cite{DBLP:journals/corr/SongYLZZ16,yang2019hybrid,cai2019skeleton,weston2018retrieve}. \textit{Thus, we propose retrieving factual information to improve a persuasive dialogue system's ability to consistently and coherently address user questions, which may lead to improved perceptions of intelligence, coherence, and trustworthiness.}

\section{Dataset}
We use the 300 annotated anonymous English conversations in the \persuasion dataset. In each conversation, one person, the ``persuader,'' tries to convince their conversational partner, the ``persuadee,'' to donate to Save the Children. The conversations last for 10 turns, and a user's utterance during a turn contains at least one sentence. Each sentence is annotated with one of several dialogue acts, including inquiries (e.g. ``Have  you donated to a charity before?'') and various persuasive appeals (e.g. ``I'll match your donation, and together we can double the amount!''). In this work, we build a system that acts as a persuader. The full list of persuader dialogue acts used is provided along with examples in Table~\ref{dialogue_acts}. 

\begin{figure*}
    \centering
    \includegraphics[trim={0 0 0 0},clip,width=16cm]{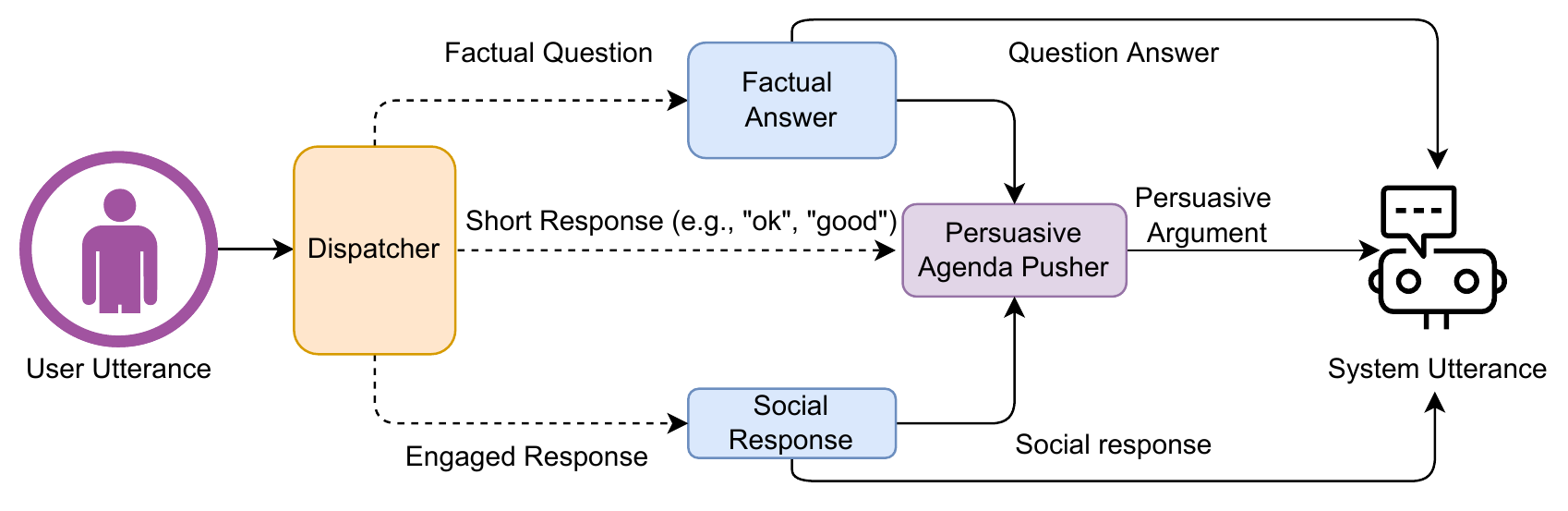}
    \caption{Overview of the \framework framework. The user's utterance is classified by the Dispatcher (orange module), which decides whether it should be sent to the Factual Answer Module, Social Response Module, or neither (blue modules). The output from this first layer is propagated into the inputs to the Persuasive Agenda Pushing Module (purple module). The outputs from the blue and purple modules are concatenated as the final system utterance.}
    \label{fig:architecture}
\end{figure*}
\section{The RAP Framework}
The dynamics of a persuasive conversation fall between that of social dialogue and task-oriented dialogue. Typically, social chatbots like Blenderbot \cite{komeili2021internet, xu2021beyond} are used to engage with users in chit-chat, and language models like BART \cite{lewis-etal-2020-bart} are used in controllable generation~\cite{wu2021controllable}. However, it is difficult for one end-to-end model to perform both tasks.
We break down the problem of generating a persuasive response into two parts: 1) generating an utterance that \textit{responds} to users' comments, questions and concerns, and 2) generating an utterance that \textit{pushes the agenda} of a conversation. In this context, pushing an agenda refers to progressing through a set of persuasive strategies as in Table~\ref{dialogue_acts}. We propose interweaving responses with agenda-pushing within the same turn, inspired by the joint goal balancing in \citet{zhang-danescu-niculescu-mizil-2020-balancing}. 
As outlined in Figure~\ref{fig:architecture}, our framework comprises four core components: a \textit{dispatcher} to decide which response modules to invoke, a \textit{factual answer} module and a \textit{social response} module to acknowledge and respond to users, and an \textit{agenda-pushing model} to ensure the persuasive conversation stays on task. 

\subsection{The Dispatcher}
Upon receiving an utterance from a user, \framework first invokes the Dispatcher to decide which response module(s) to invoke. It 
classifies the dialogue act of the user utterance using a dialogue act classifier from \citet{shi2020refine}. As shown in Figure~\ref{fig:architecture}, if the utterance includes a factual question or task-related inquiry as determined by its dialogue act or regular expressions, the Dispatcher will invoke the Factual Answer Module. If the dialogue act instead indicates that it is a statement that shows engagement\footnote{The dialogue act must not be ``acknowledgement.''} with the chatbot, the Dispatcher will invoke the Social Response Module. The output of the Factual Answer and Social Response modules is propagated to the Agenda Pusher. 

\subsection{Creating Engagement via User Response}
\paragraph{The Factual Answer Module}
In order to maintain consistency in answers, we compute the cosine distance of Sentence-BERT \cite{reimers-gurevych-2019-sentence} embeddings between the user's question and question-answer mappings from the training data. The question-answer mappings are also built using Sentence-BERT by aggregating the answers of all of the most similar questions. We retrieve the answer to the question that has the lowest cosine distance in semantic meaning from the question asked by the user.
\paragraph{The Social Response Module}
The Social Response Module comprises of a pretrained Blender Bot 2.0 instance with 3B parameters, an updated version of the open-domain BlenderBot social chatbot~\cite{roller-etal-2021-recipes}, that builds long-term memory and queries the internet\footnote{We use a publicly available implementation of Blender Bot 2.0 that makes use of a Google search retriever.}.
We feed the model a context string consisting of the conversation history and generate a response in a zero-shot setting. We do not keep outputs that Blender Bot 2.0 labels as ``potentially unsafe.'' Finally, we still want to push the agenda of the conversation, regardless of whether or not the Social Response or Factual Answer modules were invoked to generate a directed response towards the user. 
\subsection{The Persuasive Agenda-Pushing Module}
We ensure that the conversation stays on the persuasive agenda using conditional generation with BART~\cite{lewis-etal-2020-bart}\footnote{BART Large, 406M parameters; fine-tuning code given.}, a pre-trained Transformer language model.
If the Factual Answer or Social Response modules are invoked, the response is appended to the conversation history, which is included as input to BART for consistency.
\subsubsection{Conditional Generation Background}
For our agenda-pushing model, we fine-tuned BART on the Persuasion4Good dataset using HuggingFace's Transformers package~\cite{wolf-etal-2020-transformers}. However, it is not enough to just perform language modeling: \textit{an automated persuasive dialogue system should incorporate pragmatic persuasive strategies to ensure the conversation stays on task}. Thus, we draw inspiration from CTRL \cite{keskar2019ctrl}, a state-of-the-art Transformer model for conditional generation. 

Traditionally, language modeling is framed as a problem of learning next-word prediction and the objective is to minimize the negative log likelihood, $L(D)$, over a dataset $D=\{x_1, x_2, ..., x_{|D|}\}$.


However, CTRL  conditions on a control code $c$, reformulating next-word prediction as $P(x|c)$ (equation~\ref{Pxc}),

\begin{equation}
    P(x|c) = \displaystyle \prod_{i=1}^n P(x_i|x_{<i}, c)
    \label{Pxc}
\end{equation}

and reformulating the negative log likelihood conditionally (equation~\ref{Lc}).

\begin{equation}
    L_c(D) = - \displaystyle \sum_{k=1}^{|D|} \log(p_\theta(x_i^k|, x_{<i}^k, c))
    \label{Lc}
\end{equation}
\subsubsection{Conditional Generation with Pragmatic Persuasive Strategies}
\label{training}
In CTRL, the control codes were used to control aspects of language such as style and content. In our study, we create a system that conditions on pragmatic dialogue acts (e.g., persuasive strategies). The agenda of dialogue acts is listed in order in Table~\ref{dialogue_acts} along with an example of each.
This ordering was determined in \citet{wang2019persuasion} as the most probable dialogue act at each turn. 

To this end, we fine-tune BART on the Persuasion4Good dataset, randomly selecting $80\%$ of the conversations as a training set. and $10\%$ as a validation set. A design decision of note is the construction of each training instance. Since the Persuasion for Good dataset contains multiple sentences (and consequently, multiple dialogue acts) per turn, one must choose between having each training instance represent one sentence as the target utterance, or a concatenation of several sentences as the target utterance. We ultimately chose to follow the latter in order for the model to learn more coherent generation. However, multiple-sentence conditional text generation also results in a more complicated task than classic single-sentence generation tasks. 

Drawing inspiration from~\citet{li2020end}, each training instance $i$ is ultimately represented as a concatenation of the \textit{history of the persuader and persuadee utterances}, the \textit{previous dialogue act}, and the \textit{planned dialogue act} on turn $i$ (i.e., the ground-truth annotated dialogue act associated with the target utterance).

While one can train a conditional generation model according to $L_c(D)$ through methods such as concatenating control codes to the end of the input sequence, we find that on the \persuasion dataset, such models cannot learn to consistently generate utterances according to the correct dialogue act. We thus add a penalty during loss computation, resulting in $L_p(D)$ 
(equation~\ref{Lp}): 

\begin{equation}
    L_p(D) = \displaystyle L_c(D) + \alpha * [f_{dc}(y) \neq c]
    \label{Lp}
\end{equation}

where $f_{dc}(y)$ is the output of a dialogue act classifier as described in \citet{shi2020refine} (a GPT-2 based model achieving the state-of-the-art on the \persuasion task: $0.66$ F1), $y$ is the generated utterance of a model given $x_{<i}^k, c$, and $\alpha$ is a tunable penalty for generating an utterance that does not match dialogue act $c$ (i.e., when $f_{dc}(y) \neq c$). $\alpha$ is tuned throughout the training process, in addition to other hyperparameters such as the learning rate.\footnote{For each hyperparameter setting, we used a fixed decoding method --- beam sampling with n-gram blocking.}
\section{Evaluation}
We evaluate \framework against an end-to-end fine-tuned BART model as described in Section~\ref{training}. This allows us to directly evaluate the impact of integrating factual information and social content and persuasive strategies in contrast to a conversation only driven by persuasive strategies. 

We evaluate the performance of the conditional generation model by calculating the dialogue act accuracy on a withheld test set consisting of $10\%$ of all conversations. As language generation is non-deterministic, we average the dialogue act accuracy across ten passes. We chose BART over Blenderbot in the Persuasive Agenda-Pushing Module because Blenderbot did not achieve as strong of a dialogue act accuracy. This is likely because Blenderbot is better-suited for social dialogue, whereas the dialogue act utterances are largely task-oriented in nature. Additionally, we specifically do not use metrics such as perplexity to compare the BART baseline and \framework because RAP is a result of several different components, and not all of which do we train or fine-tune. Additionally, because of the penalty added in $L_p$, training perplexity is no longer interpretable. It also cannot be compared to other models in other work that has used the \persuasion dataset such as \citet{li2020end}, as the model sizes differ. Most importantly, the primary objective is to build a more persuasive dialogue system, making it imperative to emphasize users' perception and conversation experience. Thus, to compare between the two frameworks, we primarily rely on feedback from human evaluation. We additionally compare utterance-based proxies for user engagement in Table~\ref{Lengths}.

\begin{table}[]
\centering
\begin{tabular}{l|l|l}
\textbf{Utterance Statistic}               & \textbf{Baseline}  & \textbf{RAP}   \\ \hline
\# Chatbot Words     & 11.14 & 16.41 \\
\# User Words        & 3.70  & \textbf{5.75}**  \\
\# Chatbot Sentences & 1.02  & 1.48  \\
\# User Sentences    & 1.09  & \textbf{1.17}** 
\end{tabular}
\caption{Average number of words and sentences per turn for both the chatbot and the user in conversations with both the baseline (BART) and RAP. ** statistically significant differences in user reply length ($\alpha=0.05$).}
\label{Lengths}
\end{table}
\section{Experimental Setup}
We deployed our chatbot using the LegoEval platform~\cite{li-etal-2021-legoeval}. The chatbot is given a gender-neutral name, Jaime. The task consists of a pre-task survey, a conversation where each participant responds to the chatbot with a minimum of seven and maximum of ten conversational lines, and a post-task survey. The pre-task survey consists of questions about demographic information (e.g., age, gender, income) and a test of the Big Five personality traits~\cite{goldberg1992development}. The post-task survey asks participants about their conversation experiences. It includes an attention validation question ("What charity was the chatbot talking about?") then asks about the users' intention to donate to Save the Children and their perception of the chatbot, including evaluations on various traits such as perceived competence and warmth. The full lists of questions is outlined in Table~\ref{evaluation}. Each participant was asked to share their impression of the chatbot along each trait using a Likert scale. A score of 1 corresponds to ``strongly disagree'' and 5 corresponds to ``strongly agree.'' We recruited 111 students from a Natural Language Processing class at a mid-size private university in the United States in exchange for course credit. Three participants did not correctly answer the validation question, resulting in a final sample of 108 participants. Each participant interacts for seven to ten turns, resulting in a sample of up to 1080 user dialogue turns. 
We used a double-blinded, between-subjects design. Each participant was given a link that randomly assigned the participant to the chatbot running on the baseline or \frameworknospace, and completed the task once.

\begin{table*}[t]
\centering
\scalebox{0.85}{
\begin{tabular}{lllll}
\toprule
\textbf{The chatbot...}                                        & \textbf{Baseline ($\mu\pm\sigma$)} & \textbf{RAP ($\mu\pm\sigma$)} & \textbf{Invoked Social} & \textbf{Invoked Factual} \\ \midrule
is competent      $\uparrow$                                             & 2.53$\pm$0.82                          & 3.00$\pm$1.06$**$   & 2.98$\pm$1.08$**$                          & \textbf{3.03$\pm$1.00$**$}                     \\ 
is natural        $\uparrow$                                             & 2.35$\pm$1.03                          & \textbf{2.65$\pm$1.00}  & 2.65$\pm$1.04                          & 2.58$\pm$0.85                        \\ 
increased my intention to donate  $\uparrow$                             & 3.00$\pm$1.17                           & 3.19$\pm$1.13    & 3.16$\pm$1.14                          & \textbf{3.33$\pm$1.06}                    \\ 
is persuasive      $\uparrow$                                            & 2.63$\pm$1.05                           & \textbf{2.72$\pm$1.10}  & 2.65$\pm$1.12                          & 2.70$\pm$1.00                      \\ 
is well-intentioned   $\uparrow$                                         & 3.65$\pm$1.01                           & 3.84$\pm$1.01   & 3.86$\pm$1.03                          & \textbf{3.97$\pm$0.94}                     \\ 
is friendly       $\uparrow$                                             & 3.16$\pm$1.05                           & 3.39$\pm$1.12   & 3.41$\pm$1.10                          & \textbf{3.58$\pm$1.12$*$}                     \\ 
is intelligent     $\uparrow$                                            & 2.51$\pm$0.92                           & 2.74$\pm$1.07   & 2.73$\pm$1.09                          & \textbf{2.79$\pm$1.07}                    \\ 
is convincing     $\uparrow$                                             & 3.02$\pm$1.08                           & 3.11$\pm$0.89   & 3.10$\pm$0.89                          & \textbf{3.15$\pm$0.89}                     \\ 
is confident      $\uparrow$                                             & 3.35$\pm$1.01                           & 3.72$\pm$0.89$**$ & 3.71$\pm$0.91$*$                          & \textbf{3.76$\pm$0.78$*$}                        \\ 
is a strong reason for donating  $\uparrow$               & 2.67$\pm$0.92                           & \textbf{2.84$\pm$1.02}   & 2.78$\pm$1.03                          & 2.82$\pm$1.09                     \\ 
was dishonest   $\downarrow$                                               & 2.14$\pm$0.89                           & 1.91$\pm$0.80   & 1.94$\pm$0.83                          & \textbf{1.88$\pm$0.77}                     \\ 
\end{tabular}
}
\caption{Comparing mean and standard deviation of the baseline (BART) and \framework from the post-task survey. Statistically significant differences compared to the baseline at $\alpha=0.05$ are denoted with $**$; significant differences at $\alpha=0.1$ are denoted with $*$. 51 participants used the baseline and 57 participants used \framework. Of the 57 \framework participants, 51 had conversations that triggered the Social Response Module and 33 conversations triggered the Factual Answer Module. 24 conversations triggered the Social Response Module but not the Factual Answer Module, and 6 conversations triggered the Factual Answer Module but not the Social Response Module.}
\label{evaluation}
\end{table*}
\section{Results}
In this section, we discuss the results of comparing \framework and baseline only using BART, the impact of individual components of \framework, and qualitatively examine participant case studies.
\subsection{Analyzing the Impact of \framework}
Across ten passes, the BART model achieves a dialogue act accuracy of 62.38\%, and was used as a part of RAP as the Agenda-Pushing Module. In Table~\ref{Lengths}, we see that RAP yielded better engagement from the participants. On average, participants responded to \framework with 5.75 words per utterance compared to 3.70 words per utterance when responding to the baseline ($p$-value $< 0.001$). Participants were also more likely to respond to RAP with more than one sentence (average: 1.17 sentences per utterance) than the baseline (average: 1.09 sentences per utterance; $p$-value $< 0.01$). Additionally, in Table~\ref{evaluation}, we find that \framework outperforms the baseline on every single perceived trait. Most notably, we see a statistically significant difference on the competence and confidence of \framework, indicating \framework is perceived to be more capable and confident in engaging in substantial topics and persuasive contents. Beyond statistical significance, we see that \framework receives better evaluations on \textit{every} single metric in comparison to the baseline, including persuasiveness, intelligence, trustworthiness, naturalness, and increasing the user's intention to donate.
\subsection{Analyzing Individual Module Contributions}
Due to constraints on our sample size, we could not run full ablation studies where we remove individual modules of the model. Instead, we analyze the perception of \framework in conversations that invoke each of the Social and Factual Answer modules. These findings are also reported in Table~\ref{evaluation}. We additionally find that each of the Social and Factual Answer modules outperform the baseline on conversations in which they were invoked. Notably, we saw that the chatbot was perceived as friendlier and significantly more competent after invoking the Social Response module. However, while there was a difference in the perceived persuasiveness of the chatbot, the difference was much smaller. This implies that perhaps social content is less closely coupled to the persuasiveness of individual arguments. After conversations invoking the Factual Response module, we indeed see the biggest increase in perception of intelligence across all conditions, although the difference is not statistically significant. We also see the largest increase in perceptions of competence. Most surprisingly, we find the biggest increase in friendliness after conversations that invoke the Factual Answer Module. This could imply that ensuring that users' questions are answered is very important in making their voices feel heard and acknowledged.

Surprisingly, there were even modules that received statistically significant differences in ratings from the baseline even when not viewed in aggregate with \framework --- this is the case for both the Social and Factual Answer Modules on competence and confidence. The Factual Answer module also received a statistically significantly higher rating on friendliness, whereas the difference for \framework was not statistically significant. Moreover, in several cases, conversations which invoked the Factual Answer module received the best-performing scores on average. Both of these findings are likely due to the fact that in nearly all cases where the Factual Answer module was invoked, the Social Response module was also invoked, but the inverse is not true. This may also indicate that the results in the Invoked Factual column is the most holistic representation of the complete \framework framework.

\subsection{Qualitative Case Studies}
\label{qualitative}
We find that participants who actively engaged \framework were able to hold coherent, intelligent conversations. \autoref{fig:comparison} shows an example of a participant who had previously heard of Save the Children. The participant had commented on their view of the importance of Save the Children, and the chatbot running using RAP was able to acknowledge their opinion (``I agree''), while further elaborating on their discussion topic (``There is a lack of support for children ... in war zones''). This statement was used to condition the agenda-pushing emotional appeal (``It's so hard to imagine what it's like for a child to grow up facing the daily threat of violence''). The full conversation is provided in Table~\ref{example_1} in \autoref{sec:examples}. User anecdotes included mentioning that they were ``pleasantly surprised'' by the ability of \framework to acknowledge them with remarks like ``I agree.'' Two full conversations with the baseline dialogue system are also provided in Tables~\ref{example_3} and \ref{example_4}. The baseline system generally appears to perform well at generating utterances according to the right dialogue act (e.g., ``I have a great story about how I helped a child in need in the first two months of the new year'' for the ``personal story'' dialogue act in Table~\ref{example_4}). In contrast to \frameworknospace, users often quickly lose interest in the dialogue system, as they do not feel acknowledged. Participants who only interacted with the baseline complained that their questions went unanswered (e.g. User: ``Do you know who is their founder?'' Chatbot: ``They are an international NGO ...''), and thus questioned whether their input was even considered by a model. 

Despite these improvements, \framework does not seem to handle current events well. In general, conditioning on social content and factual information appears to greatly improve the quality of the Agenda-Pushing Module's generation. However, when Blender Bot 2.0 cannot generate a safe output, the Agenda-Pushing Module does not seem to handle such out-of-domain instances well. 
One participant commented on the ongoing war in Ukraine. Blender Bot 2.0 was unable to produce a safe output, leaving the Agenda-Pushing Module to come up with a relevant response. However, Ukraine never appears in the training data, so the module's conditional generation model instead mentions conflicts in several other countries, and performs self-modeling. Such behavior can come across as dismissive or tone-deaf towards the user. The full conversation is provided in Table~\ref{example_2} of \autoref{sec:examples}.
While this particular implementation of \framework leveraging Blender Bot 2.0\footnote{Recent concurrent work (Blender Bot 3.0) has examined dialogue safety with a different internet retriever.} and a fixed knowledge source for retrieval may have issues with current events, \framework is general enough that it could potentially be updated with new knowledge and improved internet retrieval modules in the future which can more consistently generate safe outputs.

\section{Discussion}
Overall, we find that \framework and each of its individual modules is able to outperform state-of-the-art conditional generation models on \persuasion. One of the core advantages of end-to-end conditional generation models is that they are easily transferrable to different datasets. But, \framework is also easily transferrable --- the only requirement is that the dataset contains a set of dialogue acts with sufficient data to train a classifier, as the biggest bottleneck is being able to use a dialogue classifier for $L_p$ and in building the Dispatcher. 
On smaller datasets, it may even be possible to perform transfer learning using a classifier pre-trained on the \persuasion dataset. The Social Response Module is directly transferrable, as we are able to achieve high quality results using it zero-shot, and the Factual Answer Module uses Sentence-BERT to group together training data.

\paragraph{Limitations}
Due to the cost of human evaluation, our sample size is relatively small, 51 and 57 people for the two conditions. This limitation restricted us from performing a full ablation in which we evaluated chatbots which used each module individually.
We hope to obtain larger samples in the future to better evaluate the efficacy of our system. 

Additionally, considering the sample consists of students enrolled in Natural Language Processing, they possess a more technical background with higher standards for chatbots than the average user on Mechanical Turk. Moreover, because the sample did not enter as participants out of personal interest in Save the Children, they are less likely to be interested in childrens' charities than an individual on the internet who goes out of their way to interact with such a chatbot, which may be reflected in evaluation scoring. Anecdotally, we see in Section~\ref{qualitative} that individuals who do have some sort of inclination towards charitable organizations are actually quite positive and receptive towards the chatbot. In this regard, we are likely limited by the funds necessary to acquire a sample whose interests better align with \textsc{PersuasionForGood}. Our work faces several challenges to ultimately evaluate the hypothesis that persuasive conversations should be handled as jointly social and task-oriented.

While the dialogue act accuracy of the Agenda-Pushing module is only $62.3\%$, this metric is bottlenecked by $f_{dc}$ in equation~\ref{Lp}; the F1-score of the classifier is only $0.66$ (the state-of-the-art on the \persuasion dataset), implicitly limiting the upper bound of any generation model that is reliant on it. We find from users' conversation experiences that the chatbot more than sufficiently presents persuasive strategies. If one has a dialogue act classifier with stronger performance, they would be able to improve the ability of their agenda-pushing model to learn persuasive strategies even further. We additionally find that \textit{without} a dialogue act classifier (i.e., without $L_p$), BART is unable to achieve a dialogue act accuracy higher than $30\%$ on the \persuasion dataset.

\section{Conclusion}
Overall, we find perceptual improvements by specifically integrating social content and factual information into persuasive dialogues with \framework compared to a strong end-to-end conditional generation model like BART. While existing methods like \citet{li2020end, wu-etal-2021-alternating} achieve strong performance on automatic metrics like perplexity, \framework directly emphasizes upon users' conversational experience with a modular design rooted in social science theory. \framework is generalizeable and may even be applied towards persuasive contexts outside of charitable conversations, e.g., in the case of therapy and crisis counseling~\cite{zhang-danescu-niculescu-mizil-2020-balancing} where there are also split goals (ensuring users feel heard and pushing a conversational agenda). Future work on persuasive dialogue systems could consider implementing a strategy planner using supervised learning. Additionally, researchers could consider looking for relationships between personality data, persuasive strategies, and persuasion outcomes.

\section{Acknowledgements}
Thanks to Intel for supporting this work through a research gift. We also thank Kun Qian and Yu Li for helpful discussions and feedback, and our reviewers for their suggestions. We are grateful to all of our study participants for their help in evaluating our systems.
\bibliography{anthology,custom}
\bibliographystyle{acl_natbib}

\clearpage

\appendix
\section{Ethical Considerations}
All participants were informed that they were talking to a chatbot developed by their university's researchers. This ensures transparency in experiment design, so that participants will never feel ambiguity or discomfort with respect to whether they are speaking with a human or a chatbot. Participants also gained additional insight about their own communication styles based on the results of their Big Five personality test. All data collection associated with this task has been declared exempt by an ethics review board. All data was collected anonymously. E-mails were voluntarily provided for credit, but stored separately from the anonymized data.

Persuasion is a tricky social dynamic. It has been heavily studied, and the intention of this work, like the \persuasion dataset used, is that persuasive dialogue systems should only ever be created for social good. All related applications discussed are intended to create good for the world at an individual and societal level.
\onecolumn
\section{Example Conversations}
\label{sec:examples}
Tables~\ref{example_1} and \ref{example_2} are example full-length conversations with chatbots using the \framework framework. 
\begin{table*}[hbt!]
\small
\begin{tabular}{p{0.1\textwidth}|p{0.8\textwidth}}
Party & Utterance \\ \hline
System & Hello, how are you today? \\
User & Hi! I'm good, how are you? \\
System & \textcolor{blue}{I'm terrific!} Have you heard of the charity Save the Children? \\
User & Yes, I have heard of them! They are one of the leading charities on health. \\
System & \textcolor{blue}{That's great!} They do a lot of great things or children in developing countries. \\
User & Yes, their work includes everything from policy to on-the-ground assistance for children. It's great to know that there are such helpful organizations who focus on child health/education. \\
System & \textcolor{blue}{That's great.} It's so sad to see how many children die from preventable causes. \\
User & I agree, a lot of developing countries don't have systems or infrastructures to protect children from these causes. \\
System & \textcolor{blue}{I agree. There is a lack of support for children in developing countries, especially in war zones.} It's so hard to imagine what it's like for a child to grow up acing the daily threat of violence. \\
User & Charities might not be able to help with things like war, so that's where charities' roles become challenging. They need to deliver essential health care and services to children or the long-term. \\
System & \textcolor{blue}{That's a really good point.} I am going to donate a bit of my money to Save the Children \\
User & That's so great! You encourage me to get into the habit of donating as well. \\
System & Every little bit helps. Even a small amount! \\
User & Yup! Thank you for doing some good for the organization and the people it serves. \\
System & Thank you so much for the kind words.
\end{tabular}
\caption{Example conversation invoking the Social Response Module but not the Factual Answer Module. Blue text: generated by Social Response Module. Black text: generated by Persuasive Agenda-Pushing Module.}
\label{example_1}
\end{table*}

\begin{table*}[!htbp]
\small
\begin{tabular}{p{0.1\linewidth}|p{0.8\linewidth}}
Party & Utterance \\ \hline
System & Hello, how are you today? \\
User & I am doing very well. Thanks for asking. \\
System & {\textcolor{blue}{Thanks for your kind words.} Do you know about Save the Children?} \\
User & I have little knowledge about this. \\
System & Save the Children is an international Non-Governmental organization that promotes children's rights, provides relief and helps support children in developing countries. \\
User & I see. \\
System & It is heartbreaking to see the suffering of children in war zones. \\
User & I agree with you. This is devastating. You could see what is happening in Ukraine right now... \\
System & \textcolor{blue}{I can't imagine what it is like to live in a war zone.} A child's life is more important than any other thing in the world . a lot of children grow up facing the daily threat of violence. \\
User & I kind of worry for those children in Ukraine. \\
System & I also worry for the children in Iraq and Afghanistan and Syria and Libya. I am going to donate a bit of my money to Save the Children. \\
User & Let me know how you would like to donate. \\
System & \textcolor{blue}{I will donate to a charity that helps Ukrainian children.} Every little bit helps. \\
User & That's good to hear. \\
System & Save the children has a huge impact on children's lives. Your donation will go a long way in helping these children. \\
User & Hmm. Okay.  I agree with you, but how do I do that? \\
System & \textcolor{orange}{You can donate directly to the charity.} Would you be interested in donating a bit of money to Save the Children? \\
User & Do you have a link or something I could directly go to? \\
System & \textcolor{orange}{I sure do as a matter of fact it's \textless{}URL\textgreater.} Sorry, I have to go now, but it's been great talking to you! Thanks for your time! \\
User & Cool. See you.
\end{tabular}
\caption{Example conversation invoking the Social Response Module and the Factual Answer Module. Blue text: generated by Social Response Module. Orange text: retrieved by Factual Answer Module. Black text: generated by Persuasive Agenda-Pushing Module.}
\label{example_2}
\end{table*}
\begin{table*}[!htbp]
\small
\begin{tabular}{p{0.1\linewidth}|p{0.8\linewidth}}
Party & Utterance \\ \hline
System & Hello. \\
User & Hi how are you \\
System & Have you heard of the charity Save the Children? \\
User & Nope, what was that? \\
System & Have you ever donated to them before? \\
User & No \\
System & They are a Non-Governmental organization that promotes children's rights, provides relief and helps support children in developing countries. \\
User & Got it \\
System & It's heartbreaking to think of all the children who have grown up facing the daily threat of violence. \\
User & Sounds good \\
System & It's so hard to imagine what it would be like to have to go through what these children have gone through \\
User & Yes, I agree \\
System & I am going to donate a bit of my money to Save the Children \\
User & Excellent! \\
System & Yes, I think it is a great idea . A little bit of money goes a long way. \\
User & You bet
\end{tabular}
\caption{Example conversation using the baseline end-to-end BART model. The user starts off relatively engaged but quickly loses interest in the conversation.}
\label{example_3}
\end{table*}
\begin{table*}[!htbp]
\small
\begin{tabular}{p{0.1\linewidth}|p{0.8\linewidth}}
Party & Utterance \\ \hline
System & Hello, how are you today? \\
User & I am well, how are you? \\
System & Have you heard of Save the Children? \\
User & No \\
System & Have you ever donated to them? \\
User & No \\
System & They are an international Non-Governmental organization that promotes children's rights, provides relief and helps support children in developing countries. \\
User & Interesting, thank you for sharing \\
System & They are suffering from the effects of war, famine, and violence. \\
User & damn \\
System & It is so easy to forget that millions of people die of hunger, poverty, and disease every year. \\
User & That's true \\
System & I am going to donate a bit of my money to Save the Children \\
User & Good for you! \\
System & Yes, it will help a lot \\
User & Hopefully others can also donate \\
System & I have a great story about how I helped a child in need in the first two months of the new year. \\
User & Cool \\ 
System & Would you be interested in donating a bit of money to Save the Children? \\
User & Maybe \\
System & I hope you have a great day! \\
User & You too!
\end{tabular}
\caption{Example conversation using the baseline end-to-end BART model. Like in Table~\ref{example_4}, the user starts off relatively engaged but quickly loses interest in the conversation. }
\label{example_4}
\end{table*}
\clearpage
\section{Experiment Instructions}
\begin{figure}[hbt!]
    \centering
    \includegraphics[width=0.9\linewidth]{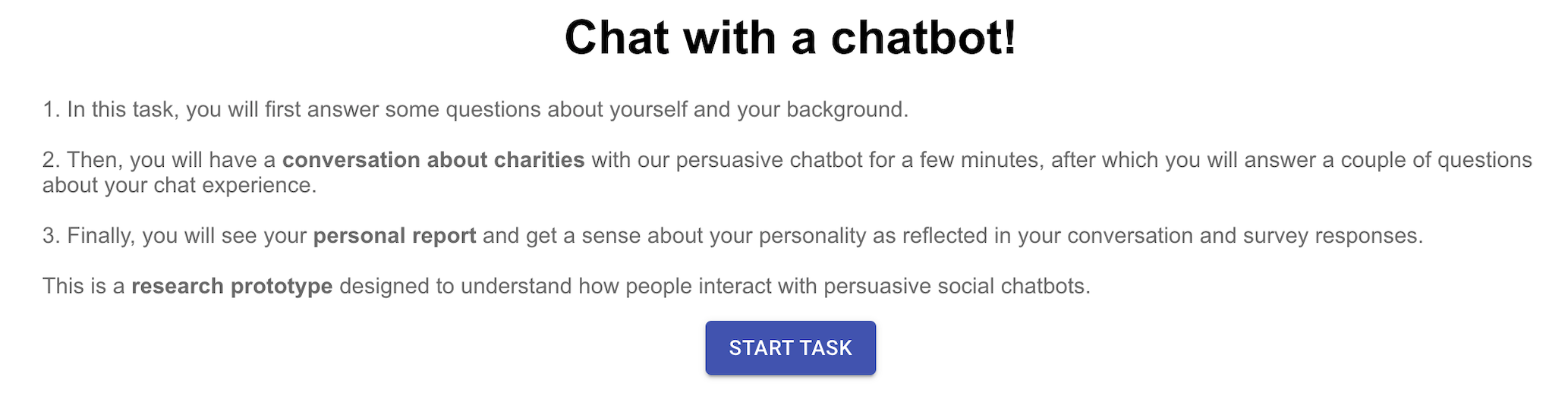}
    \caption{Screenshot of instructions given to participants. Participants are told that this chatbot is a research prototype, and are given course extra credit in exchange for participation.}
    \label{fig:landing}
\end{figure}
\begin{figure}[!htbp]
    \centering
    \includegraphics[width=0.9\linewidth]{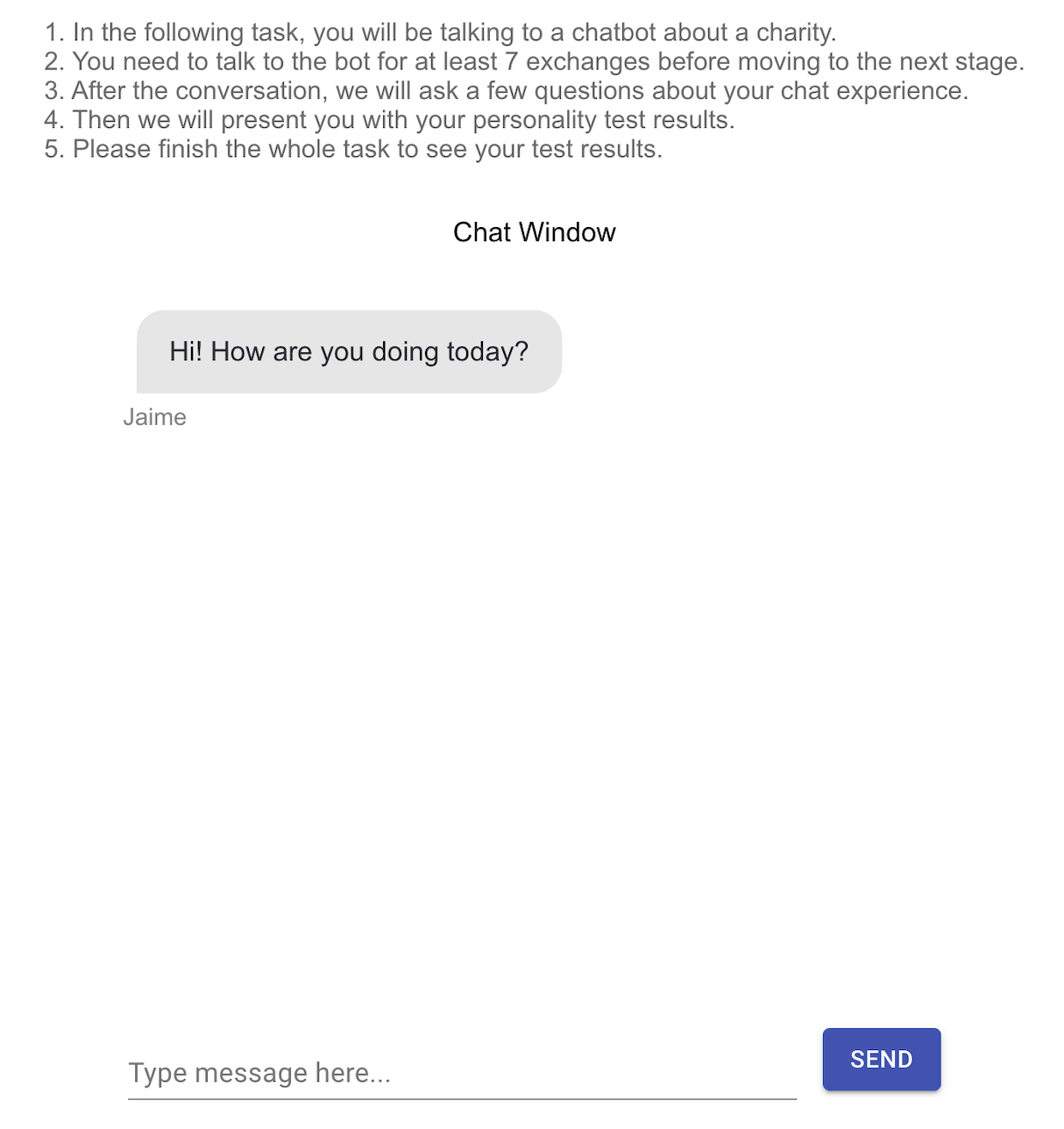}
    \caption{Screenshot of the conversation instructions and chat window presented to particpants.}
    \label{fig:chatbox}
\end{figure}
\end{document}